\documentclass[conference]{IEEEtran}
\IEEEoverridecommandlockouts
\usepackage{cite}
\usepackage{amsthm,amsmath,amssymb,amsfonts}
\usepackage{algorithm}
\usepackage{algorithmic}
\usepackage[linesnumbered,ruled,vlined,algo2e]{algorithm2e}

\usepackage{graphicx}
\graphicspath{ {./images/} }
\usepackage{textcomp}
\usepackage{xcolor}
\usepackage{caption}
\usepackage{verbatim} 
\usepackage{url}  
\usepackage{array}
\usepackage{makecell}
\usepackage{multirow}
\usepackage{enumitem}
\usepackage{booktabs} 
\usepackage{siunitx} 

\newtheorem{definition}{Definition}[section]
\NewDocumentCommand{\rot}{O{70} O{1em} m}{\makebox[#2][l]{\rotatebox{#1}{#3}}}

\def\BibTeX{{\rm B\kern-.05em{\sc i\kern-.025em b}\kern-.08em
    T\kern-.1667em\lower.7ex\hbox{E}\kern-.125emX}}

\usepackage[acronym]{glossaries}
\setacronymstyle{long-short}
\makeglossaries
\newacronym{ai}{AI}{Artificial Intelligence}
\newacronym{tl}{TL}{Transfer Learning}
\newacronym{mtl}{MTL}{Multi-Task Learning}
\newacronym{ml}{ML}{Machine Learning}
\newacronym{dl}{DL}{Deep Learning}
\newacronym{rl}{RL}{Representation Learning}
\newacronym{fl}{FL}{Feature Learning}
\newacronym{r2}{R$^2$}{Coefficient of Determination}
\newacronym{evs}{EVS}{Explained Variance Score}
\newacronym{mae}{MAE}{Mean Absolute Error}
\newacronym{mse}{MSE}{Mean Squared Error}
\newacronym{rmse}{RMSE}{Root Mean Squared Error}
\newacronym{tt}{TT}{Training Time}
\newacronym{ppe}{PPE}{Personal Protective Equipment}
\newacronym{covid19}{COVID-19}{Corona Virus Disease 2019}
\newacronym{chc}{CHC}{Community Health Centres}
\newacronym{sars-cov-2}{SARS-CoV-2}{Severe Acute Respiratory Syndrome Coronavirus 2}
\newacronym{abm}{ABM}{Agent-based Model}
\newacronym{mm}{MM}{Mathematical Model}
\newacronym{rna}{RNA}{Ribonucleic Acid}
\newacronym{uk_variant}{UK/British variant}{Lineage B.1.1.7}
\newacronym{sa_variant}{South African variant}{Lineage B.1.351}
\newacronym{brz_variant}{Brazilian variant}{Lineage P.1}
\newacronym{india_variant}{Indian variant}{Lineage B.1.617} 
\newacronym{infect}{Infected}{Infection/Positive cases}
\newacronym{hospital}{Hospitalized}{Hospitalization cases}
\newacronym{recover}{Recovered}{Recovery cases}
\newacronym{death}{Death}{Death/Mortality cases}
\newacronym{sir}{SIR}{Susceptible-Infectious-Recovered}
\newacronym{sis}{SIS}{Susceptible-Infectious-Susceptible}
\newacronym{sird}{SIRD}{Susceptible-Infectious-Recovered-Deceased}
\newacronym{msir}{MSIR}{MaternallyDerivedImmunity-Susceptible-Infectious-Recovered}
\newacronym{seir}{SEIR}{Susceptible-Exposed-Infectious-Recovered}
\newacronym{seis}{SEIS}{Susceptible-Exposed-Infectious-Susceptible}
\newacronym{suqc}{SUQC}{Susceptible-UnquarantinedInfected-QuarantinedInfected-ConfirmedInfected}

\begin{document}

\title{A \acrlong{mtl} Framework for \acrshort{covid19} Monitoring and Prediction of \acrshort{ppe} Demand in \acrlong{chc} \\
\thanks{This research was supported by the Canadian Institutes of Health Research (CIHR) Operating Grant and the Vector Institute for Artificial Intelligence.}
}

\author{
\IEEEauthorblockN{1\textsuperscript{st} Bonaventure C. Molokwu}
\IEEEauthorblockA{
\textit{School of Computer Science} \\
\textit{University of Windsor}\\
Windsor - Ontario, Canada \\
molokwub@uwindsor.ca \\
0000-0003-4370-705X
}
\and
\IEEEauthorblockN{2\textsuperscript{nd} Shaon Bhatta Shuvo}
\IEEEauthorblockA{
\textit{School of Computer Science} \\
\textit{University of Windsor}\\
Windsor - Ontario, Canada \\
shuvos@uwindsor.ca \\
0000-0002-4734-7867
}
\and
\IEEEauthorblockN{3\textsuperscript{rd} Ziad Kobti}
\IEEEauthorblockA{
\textit{School of Computer Science} \\
\textit{University of Windsor}\\
Windsor - Ontario, Canada \\
kobti@uwindsor.ca \\
0000-0001-9503-9730
}
\and
\IEEEauthorblockN{4\textsuperscript{th} Anne Snowdon}
\IEEEauthorblockA{
\textit{SCAN in Health} \\
\textit{University of Windsor}\\
Windsor - Ontario, Canada \\
anne.snowdon@uwindsor.ca
} }

\maketitle

\begin{abstract}
Currently, the world seeks to find appropriate mitigation techniques to control and prevent the spread of the new \acrshort{sars-cov-2}. In our paper herein, we present a peculiar \acrlong{mtl} framework that jointly predicts the effect of \acrshort{sars-cov-2} as well as Personal-Protective-Equipment consumption in \acrlong{chc} with respect to a given socially interacting populace. Predicting the effect of the virus (\acrshort{sars-cov-2}), via studies and analyses, enables us to understand the nature of \acrshort{sars-cov-2} with reference to factors that promote its growth and spread. Therefore, these foster widespread awareness; and the populace can become more proactive and cautious so as to mitigate the spread of \gls{covid19}. Furthermore, understanding and predicting the demand for \acrlong{ppe} promotes the efficiency and safety of healthcare workers in \acrlong{chc}. Owing to the novel nature and strains of \acrshort{sars-cov-2}, relatively few literature and research exist in this regard. These existing literature have attempted to solve the problem statement(s) using either \acrlong{abm}s, \acrlong{ml} Models, or \acrlong{mm}s. In view of this, our work herein adds to existing literature via modeling our problem statements as \acrlong{mtl} problems. Results from our research indicate that government actions and human factors are the most significant determinants that influence the spread of \acrshort{sars-cov-2}.
\end{abstract}


\section{Introduction} \label{intro}
The \gls{sars-cov-2} is a new form of enveloped \gls{rna} virus, responsible for the \gls{covid19} pandemic, and it has never been witnessed before December 2019. \gls{covid19} global epidemiological summaries \cite{WHOCD2020}, as at February 2021, report that there have been over $ 110,384,747 $ confirmed/positive cases as well as $ 2,446,008 $ deaths as a result of \gls{sars-cov-2}. Moreover, these global epidemiological summaries are quite overwhelming. Therefore, the \gls{covid19} pandemic requires urgent and crucial input as well as measures from every domain. From a Data Science perspective, it is noteworthy that the world can be conceptualized as one big data problem. With respect to the $ 21^{st} $ century, data essentially influences and controls virtually everything that we tend to accomplish. For us, as a human race, to successfully conquer the \gls{covid19} pandemic; we require sufficient knowledge about the \gls{sars-cov-2}. Thus, the quest for knowledge associated with the \gls{sars-cov-2} can only be acquired from information available about the virus. In turn, information is valuable resource which we can only extract from data. For that reason, there lies the importance of data and Data Science with respect to combating the \gls{covid19} pandemic.

Our work herein concentrates and proposes solutions to the following problem statements, namely:
\begin{enumerate}
\item[i.] Monitor the effect of \gls{sars-cov-2} via effective predictions of estimates with respect to \gls{infect}, \gls{hospital}, \gls{recover}, and \gls{death}.
\item[ii.] Predict the demand of \gls{ppe} by \gls{chc} that provide medical treatment to \gls{covid19} (\gls{hospital}) patients.
\end{enumerate}
With respect to our experiments and results, we have used the Canadian province of Ontario as our case study. 

At the moment, there exist four (4) prevalent variants/strains \cite{WHOCD2020} of \gls{sars-cov-2} around the world. These strains are, namely: \gls{uk_variant}, \gls{sa_variant}, \gls{brz_variant}, and \gls{india_variant}. Our research contributions and novelties herein cannot be downplayed considering the potential consequences of these emerging strains, and the current global impact of \gls{covid19}. From one viewpoint, we have been able to identify several factors (Biological, Environmental, Government Actions, and Human), which influence the spread of \gls{sars-cov-2}, by means of studies, experiments, and analyses. Also, we have been able to effectively forecast estimates of \gls{covid19} impact with respect to \gls{infect}, \gls{hospital}, \gls{recover}, and \gls{death} cases. Consequently, these contributions can serve as control as well as preventive measures in curtailing the spread of \gls{sars-cov-2}. Similarly, these contributions can be employed in clinical research and trials, with respect to drug and vaccine development, against \gls{sars-cov-2}. From another viewpoint, we have proposed a model that forecasts \gls{ppe} demand in \gls{chc} with regard to the \gls{covid19} pandemic. Considering the safety of our frontline healthcare workers, this contribution serves as a preventive and control measure toward their protection. Also, this can serve as a proactive measure toward ensuring a robust supply network of \gls{ppe} to \gls{chc}.

Furthermore, the (two) aforementioned problem statements have been modeled and analyzed as regression-based problems. Hence, we have employed a \gls{mtl} methodology with reference to the development and implementation of our proposed framework. Our parallel \gls{tl} model tackles the problem statements herein via pre-training on datasets, which are comprised of $ i \times 13 $ feature vector (row vector), collected from six (6) distinct Canadian provinces, viz: Alberta, British Columbia, Manitoba, New Brunswick, Quebec, and Saskatchewan. Subsequently, the resultant pre-trained model is referenced, as a source point for the transfer of learning and knowledge, for training another (dedicated) model concerned with the resolution of problems relating to our case study (Ontario province). 

Our research introduces the following novelties, viz:
\begin{enumerate} \label{novelties}
\item[(1)] The parallel transfer of learning from correlated (source) domains to a target domain, via pre-training a \gls{mtl} framework, yields much better generalization results with respect to \gls{covid19} monitoring.
\item[(2)] Identification of influential factors, based on four (4) categories (\gls{sars-cov-2} Biological Factors, Environmental Factors, Government Actions, and Human Factors), which affect the spread of \gls{covid19}.
\item[(3)] Human factors (precisely age-group stratification) most significantly influence the rates of \gls{infect} cases, \gls{hospital} cases, and \gls{death} cases.
\item[(4)] Government Actions (precisely vaccination, pandemic wave) most significantly influence \gls{recover} cases rate.
\item[(5)] We identified that \gls{covid19} is most prevalent among males and females, within the age group of 0 to 34, in a given socially interacting populace (see Figure \ref{Figure:infect_age_group}).
\item[(6)] Proposition of a model for the prediction of \gls{ppe} demand (by \gls{chc}) with regard to \gls{covid19} \gls{hospital} cases.
\item[(7)] Detailed evaluation and performance reports based on classic \gls{ml} objective functions.
\end{enumerate}

Our work presented hereafter is organized as follows: section \ref{lit_review} reviews a selected list of related literature. Section \ref{methodology} formally defines the problem statement as well as the details of our proposed framework. Section \ref{materials_methods} expatiates on the datasets and materials used to facilitate our experiments. Section \ref{discussions} documents the detailed results of our benchmark experiments; also, it captures our analyses, discussions, and conclusion.

\section{Historical Foundation and Related Literature} \label{lit_review}
Several literature and published work, which aimed at resolving problems related to epidemiology, can be classified into three (3) broad categories, namely: Conceptual Models, Compartmental Models, and Computational Models. 
\subsection{Conceptual Models}
Models in this category are essentially high-level representations, based on abstract ideas and notions, which illustrate how these models operate with regard to resolving targeted problems in epidemiology. A common shortcoming of these models is that they are simplistic, abstract, and usually not empirical. \cite{BenShlomo2002}, \cite{Burr2016}, \cite{Bernasconi2020}, and \cite{Lin2020ACM} primarily employed conceptual modeling toward resolving research problems in epidemiology.
\subsection{Compartmental Models}
Basically, these models are \acrlong{mm}s which are based on a series of mathematical equations. They are employed in studying and analyzing how infectious diseases spread and affect different compartments of a given socially interacting populace. Also, they have been used to forecast the potential outcomes of endemics, epidemics, and pandemics. One drawback of these models is that some of them tend to be relatively complex. Common models in this category include, namely: \gls{sir} model \cite{Brauer2008Compartmentals}, \gls{sird} model \cite{Brauer2008Compartmentals}, \gls{sis} model \cite{Brauer2008Compartmentals}, \gls{msir} model \cite{Hethcote2000}, \gls{seir} model \cite{Hethcote2000}, \gls{seis} model \cite{Hethcote2000}, \gls{suqc} model \cite{Zhao2020Modeling}, etc.
\subsection{Computational Models}
In this category, there exist two (2) major subcategories of models used for epidemiology-related problems, viz: \acrlong{abm}s and \acrlong{ml} models.

On one hand, an \gls{abm} is a computational model that re-creates a system, via simultaneously simulating the interactions of several autonomous agents, with the goal of analyzing and predicting potential event(s) about the given system. A popular downside of these models is that they tend to oversimplify, thereby yielding unauthentic predictions. Research which employed \gls{abm} for \gls{covid19} related problems include, viz: \cite{Aleta2020}, \cite{Rockett2020}, \cite{SILVA2020110088}, \cite{Shuvo2020S}, etc.

On the other hand, a \gls{ml} model is an \gls{ai} approach such that a computational model is constructed, via learning from sample (or training) data, so as to extract inherent patterns about a given system which will be applied in making predictions/decisions about the given system. A common challenge with employing these models is the availability of good and sufficient sample data for training these models. Literature which have employed \gls{ml} approach toward resolving \gls{covid19}-related problems include: \cite{Zou2020}, \cite{Kukar2020}, etc.

\section{Proposed Framework and Methodology} \label{methodology}
This section is subdivided into three (3) subsections, namely: subsection \ref{problem_def} (problem definition), subsection \ref{prop_methods} (proposed methodology), and subsection \ref{framework_algo} (proposed system framework and algorithms).

\subsection{Definition of Problem} \label{problem_def}
\begin{definition}{\gls{covid19} Monitoring:}
Given a set of feature (independent) variables, $ X \in \mathbb{R}: x_{i,1}, x_{i,2}, ..., x_{i,j} $, such that the shape of the feature space is an $ i \times j $ feature vector; and a set of target (dependent) variables, $ Y \in \mathbb{Z}: y_{i,1}, y_{i,2}, ..., y_{i,k} $, such that the shape of the target space is an $ i \times k $ target vector. Our \gls{covid19} Monitoring framework aims at training a \gls{ml} function, $ f_m : X \rightarrow Y \equiv x_{i,*} \mapsto y_{i,*} $, which learns to effectively make predictions about $ Y $ based on the patterns of information learnt from $ X $. Thus, $ y_{i,*} \in Y = f_m(x_{i,*} \in X) $.
\end{definition}

\begin{table}[h!]  
\centering
\setlength{\tabcolsep}{0.2em}  
\renewcommand{\arraystretch}{1.25}  
\linespread{0.85}  
\rmfamily  
\caption{Primary features constituting the feature space of our framework \cite{Berry2020, GoogleMobility2021, StatCan2021}.}
	\begin{tabular}{|p{5.0em}|p{2.9em}|p{5.8em}|p{15.8em}|}  
    \hline
		\multicolumn{4}{|c|}{Initial (or Primary) Features} \\
        \multicolumn{1}{|l|}{Category} & \multicolumn{1}{l|}{Code} & \multicolumn{1}{l|}{Feature Name} & \multicolumn{1}{l|}{Description or Details of Feature} \\
 	\hline
		Biological Factors & feat\_01 & Virus Reprod. Index & The Effective Reproduction Number of \gls{sars-cov-2}. \\
	\hline
		\multirow{3}{5.5em}{Environment Factors} & feat\_02 & Climate & Canadian seasonal periods of the year (1 = Spring, 2 = Summer, 3 = Autumn, 4 = Winter). \\
		& feat\_03 & Dry Land & Area (in km$^2$) of land inhabited by the populace, exclusive of aquatic habitat. \\
		& feat\_04 & Region & Numeric encoding of each region/province (0 = Alberta, 1 = British Columbia, 2 = Manitoba, 3 = New Brunswick, 4 = Newfoundland and Labrador, 5 = Nova Scotia, 6 = Ontario, 7 = Prince Edward Island, 8 = Quebec, 9 = Saskatchewan). \\
	\hline
		\multirow{7}{5.5em}{Government Actions} & feat\_05 & Wave & Pandemic phase (1 = first wave, 2 = second wave). \\
		& feat\_06 & Cumm. Vaccine & Cumulative record of inoculated persons. \\
		& feat\_07 & Lockdown & Stages of restrictions with regard to public health safety (1 = Lockdown scenario, 2 = Partial/Restricted reopening, 3 = Total/Full reopening). \\
		& feat\_08 & Travel Restrict & Implementation of travel restrictions (0 = No restriction, 1 = Federal government restriction, 2 = Provincial government restriction). \\
		& feat\_09 & Province FaceCover & Implementation of compulsory face covering (0 = Not compulsory, 1 = Mandatory/Compulsory). \\
		& feat\_10 & Holiday & Effective days of holiday (0 = Workday, 1 = Holiday). \\
		& feat\_11 & CHCentres & Total count of Community Health Centres in that region/province. \\
	\hline
		\multirow{4}{5.5em}{Human Factors} & feat\_12 & retail and recreation change & Deviation from the normal regarding visitations to/fro retail and recreation centres. \\
		& feat\_13 & grocery and pharmacy change & Deviation from the normal regarding visitations to/fro grocery and pharmacy centres. \\
		& feat\_14 & parks change & Deviation from the normal regarding visitations to/fro camps and parks. \\
		& feat\_15 & transit stations change & Deviation from the normal regarding visitations to/fro public transit stations. \\
		& feat\_16 & workplaces change & Deviation from the normal regarding visitations to/fro offices \& workplaces. \\		
		& feat\_17 & residential change & Deviation from the normal regarding visitations to/fro residential apartments and living homes. \\
		& feat\_18 & Return Travellers & Number of travelers returning to this region as their destination. \\
		& feat\_19 & Employ Rate & Employment rate (\%) of the region or province. \\
		& feat\_20 & Unemploy Rate & Unemployment rate (\%) of the region or province. \\
		& feat\_21 & Labor Popln & Eligible workforce for the region or province. \\
		& feat\_22 & 0 - 34 (M) & Male populace of age range: 0 - 34. \\
		& feat\_23 & 35 - 69 (M) & Male populace of age range: 35 - 69. \\
		& feat\_24 & 70 - Above (M) & Male populace of age range: 70 and above. \\
		& feat\_25 & 0 - 34 (F) & Female populace of age range: 0 - 34. \\
		& feat\_26 & 35 - 69 (F) & Female populace of age range: 35 - 69. \\
		& feat\_27 & 70 - Above (F) & Female populace of age range: 70 and above. \\
	\hline
    \end{tabular}
\label{Table:primary_feats_space}
\end{table}

\subsection{Proposed Methodology} \label{prop_methods}

\subsubsection{Feature Engineering Layer} \label{layer_1}
Taking into consideration a socially interacting populace; the feature space of our \gls{mtl} framework is established with respect to four (4) categories of influential factors, namely: \gls{sars-cov-2} Biological Factors, Environmental Factors, Government Actions, and Human Factors. These factors tend to affect the spread of \gls{covid19} with reference to a given populace. Features have been aggregated, based on the aforementioned categories of influential factors, with reference to any given population. Therefore, the initial or primary feature space is an $ i \times 27 $ feature vector with an elastic sample span. Table \ref{Table:primary_feats_space} provides detailed description(s) of each constituent feature for the $ i \times 27 $ feature vector.

\subsubsection{Feature Extraction Layer} \label{layer_2}
Furthermore, based on a basic examination of the initial/primary feature space, we were able to extract derived/secondary features. These derived features were computed via the application of arithmetic ratios and proportions to selected features of the initial/primary feature space. Therefore, the shape of the derived/secondary feature space is an $ i \times 17 $ feature vector with an elastic sample span. Also, the linear concatenation of the initial (primary) feature space, $ i \times 27 $, and the derived (secondary) feature spaces, $ i \times 17 $, temporarily expands our overall feature space to an $ i \times 44 $ feature vector.

\subsubsection{Feature Selection Layer} \label{layer_3}
In this layer, the dimensionality of the overall feature space, $ i \times 44 $ feature vector, is effectively and efficiently reduced to yield a feature space comprising only highly relevant features. These relevant features possess a high-degree influence with respect to the prediction of the target (or dependent) variables. Considering our study herein and experiment framework, on one hand, the shape of the final feature space of our model is an $ i \times 13 $ feature vector. This means that our proposed model learns to generalize based only on $ 13 $ highly relevant features per dataset.

\begin{table}[h!]  
\centering
\setlength{\tabcolsep}{0.25em}  
\renewcommand{\arraystretch}{1.22}  
\linespread{0.85}  
\rmfamily  
\caption{Highly relevant features constituting the final feature space of our \gls{mtl} framework.}
	\begin{tabular}{|r|p{3.6em}|p{3.0em}|p{5.8em}|r|r|r|r|}  
    \hline
		\multirow{2}{*}{} & \multicolumn{7}{c|}{Final (or Highly Relevant) Features} \\
        & \multirow{2}{3.6em}{Category} & \multirow{2}{3.0em}{Code} & \multirow{2}{5.8em}{Feature Name} & \multicolumn{4}{c|}{Relevance Score per Target Variable} \\
        & & & & \multicolumn{1}{l|}{\gls{infect}} & \multicolumn{1}{l|}{Hospital.} & \multicolumn{1}{l|}{Recover.} & \multicolumn{1}{l|}{\gls{death}} \\
 	\hline
		$ 1 $ & \multirow{13}{3.6em}{Relevant Features or Factors} & feat\_05 & Wave & 100\% & 31\% & 83\% & 18\% \\
		$ 2 $ &  & feat\_23 & 35 - 69 (M) & 85\% & 98\% & 68\% & 95\% \\
		$ 3 $ &  & feat\_21 & Labor Popln & 83\% & 93\% & 66\% & 90\% \\
		$ 4 $ &  & feat\_24 & 70 - Above (M) & 83\% & 100\% & 66\% & 100\% \\
		$ 5 $ &  & feat\_27 & 70 - Above (F) & 82\% & 99\% & 66\% & 99\% \\
		$ 6 $ &  & feat\_26 & 35 - 69 (F) & 82\% & 93\% & 66\% & 90\% \\
		$ 7 $ &  & feat\_22 & 0 - 34 (M) & 81\% & 88\% & 65\% & 84\% \\
		$ 8 $ &  & feat\_25 & 0 - 34 (F) & 81\% & 88\% & 65\% & 84\% \\
		$ 9 $ &  & feat\_11 & CHCentres & 76\% & 80\% & 62\% & 72\% \\
		$ 10 $ &  & feat\_06 & Cumm. Vaccine & 68\% & 54\% & 100\% & 42\% \\
		$ 11 $ &  & feat\_03 & Dry Land & 64\% & 92\% & 54\% & 95\% \\
		$ 12 $ &  & feat\_17 & residential change & 51\% & 60\% & 52\% & 60\% \\
		$ 13 $ &  & feat\_07 & Lockdown & 23\% & 44\% & 19\% & 52\% \\
	\hline
    \end{tabular}
\label{Table:relv_feats_space}
\end{table}

\begin{figure}[h!]  
\centering
\includegraphics[width=0.495\textwidth]{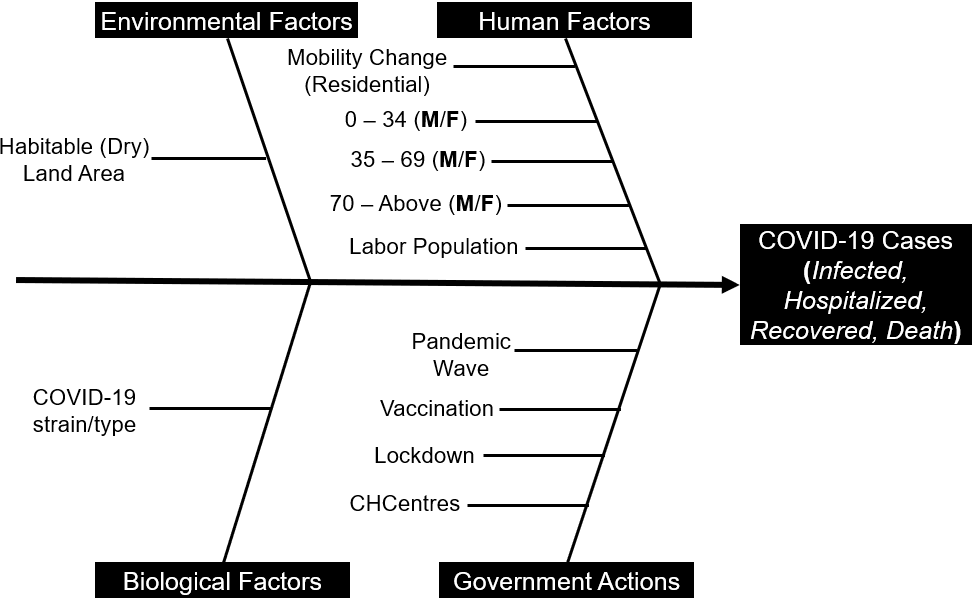}
\caption{Fishbone diagram of the relevant features affecting COVID-19 cases (case study = Ontario, Canada).}
\label{Figure:fishbone_diagram}
\end{figure}

On the other hand, the target variables comprise, viz:
\begin{enumerate} \label{target_variables}
\item[1.] \gls{sars-cov-2} infection predictions (\gls{infect}: $ y_{*,1} \subseteq Y^I $).
\item[2.] Predictions of hospitalized \gls{covid19} patients (\gls{hospital}: $ y_{*,2} \subseteq Y^H $).
\item[3.] Predictions of patients' recoveries from \gls{covid19} (\gls{recover}: $ y_{*,3} \subseteq Y^R $).
\item[4.] \gls{covid19} related death predictions (\gls{death}: $ y_{*,4} \subseteq Y^D $).
\end{enumerate} 

Table \ref{Table:relv_feats_space} and Figure \ref{Figure:fishbone_diagram} provide in detail the $ 13 $ highly relevant features which constitute the final feature space. The relevance score with respect to each target or dependent variable is indicated via columns: `\gls{infect}', `Hospital.', `Recover.', and `\gls{death}', respectively.

\subsubsection{Feature Scaling Layer} \label{layer_4}
After reviewing the final feature/independent variables of our data distribution, we noticed a lot of skewness in the data representation of the features. This problem of skewness, within the feature space, has to be overcome so as to improve the effectiveness of our model.

The constituent data of every feature variable, in the feature space, has been standardized (column-wise) to a standard normal data-distribution by means of Nonlinear Data Transformation \cite{scikit-learn} techniques as expressed in equation \ref{Equ:quant_trans}.
\begin{equation} \label{Equ:quant_trans}
\begin{aligned}
F(x_{*,j}) &\equiv \mathbb{P}(x_{*,j} \leq X) = p_{*,j} \in [0, 1], \: X \in \mathbb{R} \\
q_{*,j} \in Q &\equiv F^{-1}(p_{*,j}) = min\{x_{*,j} \in \mathbb{R}: F(x_{*,j}) \geq p_{*,j}\} \\
z_{*,j} \in Z &= \dfrac{q_{*,j} - \mu}{\sigma} \hspace{1.4cm} \text{Standard Score (Z) function} \\
\end{aligned}
\end{equation}
In equation \ref{Equ:quant_trans}, $ F $ and $ F^{-1} $ denote the Cumulative Distribution and Quantile functions, respectively. Consequently, the output ($ Q $) of the Quantile function is centered, on a mean ($ \mu = 0.0 $) and a standard deviation ($ \sigma = 1.0 $), to yield a standard normal distribution ($ Z $). Thereafter, each standard-normal row/sample, $ z_{i,*} \in Z $, of the feature space has been normalized (row-wise) to yield a unit vector via L2-Normalization \cite{AlanJeff2007} technique as denoted in equation \ref{Equ:unit_vect}. $ i $ and $ j $ denote the dimensions of the rows and columns per feature vector.
\begin{equation} \label{Equ:unit_vect}
\begin{aligned}
\hat{z}_{i,*} \in \mathbb{R} &\equiv \sum_{a=1}^{j} (z_{i,a})^2 = (z_{i,1})^2 + (z_{i,2})^2 + ... + (z_{i,j})^2 = 1 \\
\end{aligned}
\end{equation}

Taking the dependent variables into consideration, the constituents of the target space have been transformed (column-wise) to a real distribution, $ 0 \leq \mathbb{R} \leq 1 $, by means of MinMaxScaler \cite{scikit-learn} technique as expressed in equation \ref{Equ:minmax_scale}.
\begin{equation} \label{Equ:minmax_scale}
\begin{aligned}
y_{*,k}^\prime \in Y^\prime &\equiv G(Y) = \dfrac{y_{*,k} - min(y_{*,k})}{max(y_{*,k}) - min(y_{*,k})}, \: Y \in \mathbb{Z} \\
\end{aligned}
\end{equation}
$ i $ and $ k $ denote the dimensions of the rows and columns per target vector.

\subsection{Proposed System Architecture and Algorithms} \label{framework_algo}
\begin{figure}[h!]  
\centering
\includegraphics[width=0.49\textwidth]{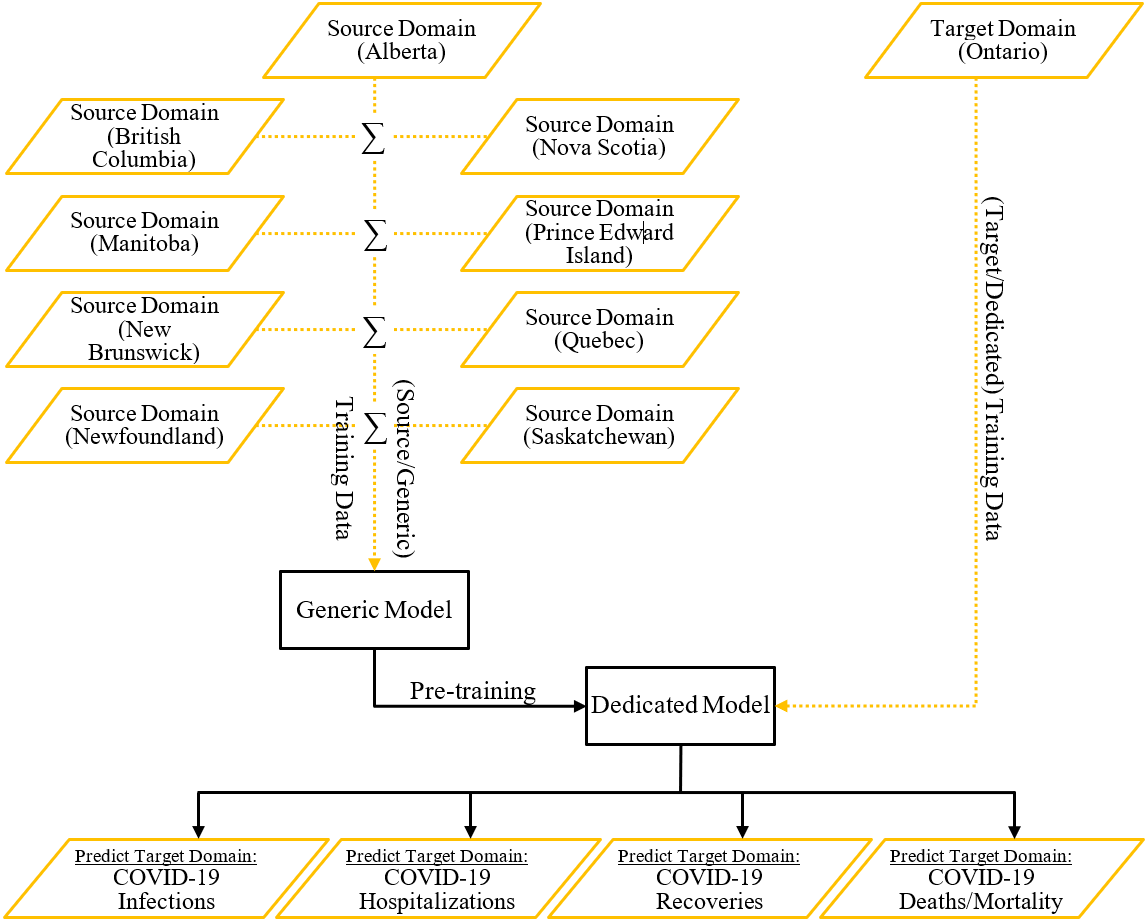}
\caption{Proposed \gls{mtl} framework for \gls{covid19} monitoring.}
\label{Figure:prop_tl_model}
\end{figure}

Training a \acrlong{ml} model solely on \gls{covid19} daily records, which are based on one regional or provincial dataset, has a great tendency for overfitting on the training dataset and/or underfitting on the validation and test datasets. At the moment, \gls{covid19} daily records spans approximately $ 450 $ records (that is $ 1 $ record per day). Thus, a training dataset comprising barely $ 450 $ records tend to yield a relatively low degree of freedom, with respect to the feature space or independent variables, during \gls{ml} training. In a bid to overcome these aforementioned challenges, we have adopted a \acrlong{mtl} technique, as represented via Figure \ref{Figure:prop_tl_model} and Algorithm \ref{alg:prop_tl_model}, to effectively improve the generalization results with respect to \gls{covid19} monitoring.

On one hand, we have trained a \emph{Generic} \gls{ml}-model component on datasets aggregated from several provinces in Canada (exclusive of the province referenced as the case study). On the other hand, we have pre-trained a \emph{Dedicated} \gls{ml}-model component via \emph{Transfer of Learning} from the \emph{Generic} \gls{ml}-model component. Subsequently, the \emph{Dedicated} \gls{ml}-model component is further trained using datasets acquired from the case study province or region.

Generalizations, with regard to predictions for the case study province, are effectuated via the \emph{Dedicated} \gls{ml}-model component of our \gls{mtl} framework. Thus, the high point of our proposed \gls{mtl} framework is that it can be readily adapted for making generalizations about any province/region. This is achieved by simply interchanging the regional dataset used for training the \emph{Dedicated} \gls{ml}-model component with a regional dataset used for training the \emph{Generic} \gls{ml}-model component.

\begin{algorithm}[H]
\caption{\acrlong{mtl} for \gls{covid19} Monitoring} \label{alg:prop_tl_model}
\SetKwInput{KwInput}{Input}                
\SetKwInput{KwOutput}{Output}              
\SetKwInput{KwData}{Data}                  
\DontPrintSemicolon
  
  \tcc{See Table \ref{Table:relv_feats_space}}
  \KwInput{$ \{X: x_{*,1}, x_{*,2}, ..., x_{*,13}\} \equiv \{X_{*,*}^{Generic}, X_{*,*}^{Dedicated} \} $}
  \tcc{See subsection \ref{target_variables}}
  \KwOutput{$ \{Y: y_{*,1}, y_{*,2}, y_{*,3}, y_{*,4}\} \equiv \{Y_{*,*}^{Generic}, Y_{*,*}^{Dedicated} \} $}
  \KwData{Regional datasets for $ X $ and $ Y $ \tcp*{See Table \ref{Table:datasets}}}

  \SetKwFunction{FMain}{Main}
  \SetKwFunction{FSub}{Sub}

  \SetKwProg{Fn}{Program}{:}{}
  \Fn{\FMain{$ X, Y $}}{
  		\tcc{Scaling: See subsection \ref{layer_4}}
  		\For{j = 1 to 13}
        {
  			$ F(x_{*,j} \in X) \equiv \mathbb{P}(x_{*,j} \leq X) = p_{*,j} $ \\
			$ q_{*,j} \in Q = min\{x_{*,j} \in \mathbb{R}: F(x_{*,j}) \geq p_{*,j}\} $ \\
			$ z_{*,j} \in Z = \dfrac{q_{*,j} - \mu}{\sigma}, \:\: \mu = 0.0, \: \sigma = 1.0 $ \\
		}
		$ \hat{z}_{i,*} \in \hat{Z} \in \mathbb{R} \equiv \sum_{a=1}^{j} (z_{i,a})^2 = 1 $ \\
		$ \hat{Z}_{*,*}^{Generic} : \hat{Z} \longrightarrow X_{*,*}^{Generic} $ \\
		$ \hat{Z}_{*,*}^{Dedicated} : \hat{Z} \longrightarrow X_{*,*}^{Dedicated} $ \\
		
		\For{k = 1 to 4}
        {
        	$ y_{*,k}^\prime \in Y^\prime = \dfrac{y_{*,k} - min(y_{*,k})}{max(y_{*,k}) - min(y_{*,k})}, \:\: Y \in \mathbb{Z} $ \\
        }
		$ Y_{*,*}^{\prime Generic} : Y^\prime \longrightarrow Y_{*,*}^{Generic} $ \\
		$ Y_{*,*}^{\prime Dedicated} : Y^\prime \longrightarrow Y_{*,*}^{Dedicated} $ \\
        
        \tcc{\acrlong{mtl}: See Figure \ref{Figure:prop_tl_model}}
        $ f_{Generic} : \hat{Z}_{*,*}^{Generic} \longrightarrow Y_{*,*}^{\prime Generic} $ \\ 
        $ f_{Dedicated} = f_{Dedicated} + f_{Generic} $ \\
        $ f_{Dedicated} : \hat{Z}_{*,*}^{Dedicated} \longrightarrow Y_{*,*}^{\prime Dedicated} $ \\
        $ Y_{i,*}^{\prime Dedicated} = f_{Dedicated}(\hat{Z}_{i,*}^{Dedicated}) $ \\
        $ Y_{i,*}^{\prime Ded.} = \{y_{i,1} \in Y^I, y_{i,2} \in Y^H, y_{i,3} \in Y^R, y_{i,4} \in Y^D\} $ \\
        
        \KwRet $ Y_{i,*}^{\prime Ded.} $
  }
\end{algorithm}

\begin{algorithm}[h!]
\caption{\gls{ppe} Consumption/Demand Prediction in \gls{chc}s} \label{alg:ppe_model}
\SetKwInput{KwInput}{Input}                
\SetKwInput{KwOutput}{Output}              
\DontPrintSemicolon
  
  \tcc{See Figure \ref{Figure:ppe_model}}
  \KwInput{$ \{y_{*,2} \subseteq Y^H, feat\_11, OprCap, Personnel \} $}
  \tcc{See Figure \ref{Figure:ppe_model}}
  \KwOutput{$ \mid PPE\,Kits_* \mid $}

  \SetKwFunction{FMain}{Main}
  \SetKwFunction{FPPEpred}{ppeDemandPred}
  \SetKwFunction{FSub}{Sub}
 
  \SetKwProg{Fn}{Function}{:}{}
  \Fn{\FPPEpred{$ Y^H, feat\_11, OprCap, Personnel $}}{
		\tcc{Initialize: Variables/Parameters}
		$ y_{i,2} \in Y^H \in \mathbb{Z} $ \tcp*{Hospitalized patients}
        $ feat\_11_i = len(CHCentres_i) $ \tcp*{Regional \gls{chc}s}
        $ OprCap_i = 0 \leq \mathbb{R} \leq 1 $ \tcp*{Operating capacity}
        \tcc{Available \gls{covid19} workforce}
        $ Personnel_i = MedLabs_i + ParaMeds_i + DoctAssts_i + Docts_i + Nurses_i + RespThpts_i $ \\
        
        $ PPE\,Kits_i = 0 $ \tcp*{Predicted \gls{ppe} kit(s)}               
        \For{i = 0 to m}
        {
            $ HspRtio_i = \dfrac{y_{i,2}}{feat\_11_i} $ \tcp*{Compute ratio}
        	\If{$ HspRtio_i > 1.0 $}
        	{
        		$ PPE\,Kits_i = OprCap_i \times Personnel_i \times 1.0 $
        	}
        	\Else
        	{
        		$ PPE\,Kits_i = OprCap_i \times Personnel_i \times HspRtio_i $
        	}
        }
        \vspace{0.3cm}
        \tcc{$ PPE\,Kits_i \approxeq i^{th} $ day \gls{ppe}-Kit demand}
        \KwRet $ PPE\,Kits_* $
  }
\end{algorithm}

\begin{figure}[h!]  
\centering
\includegraphics[width=0.49\textwidth]{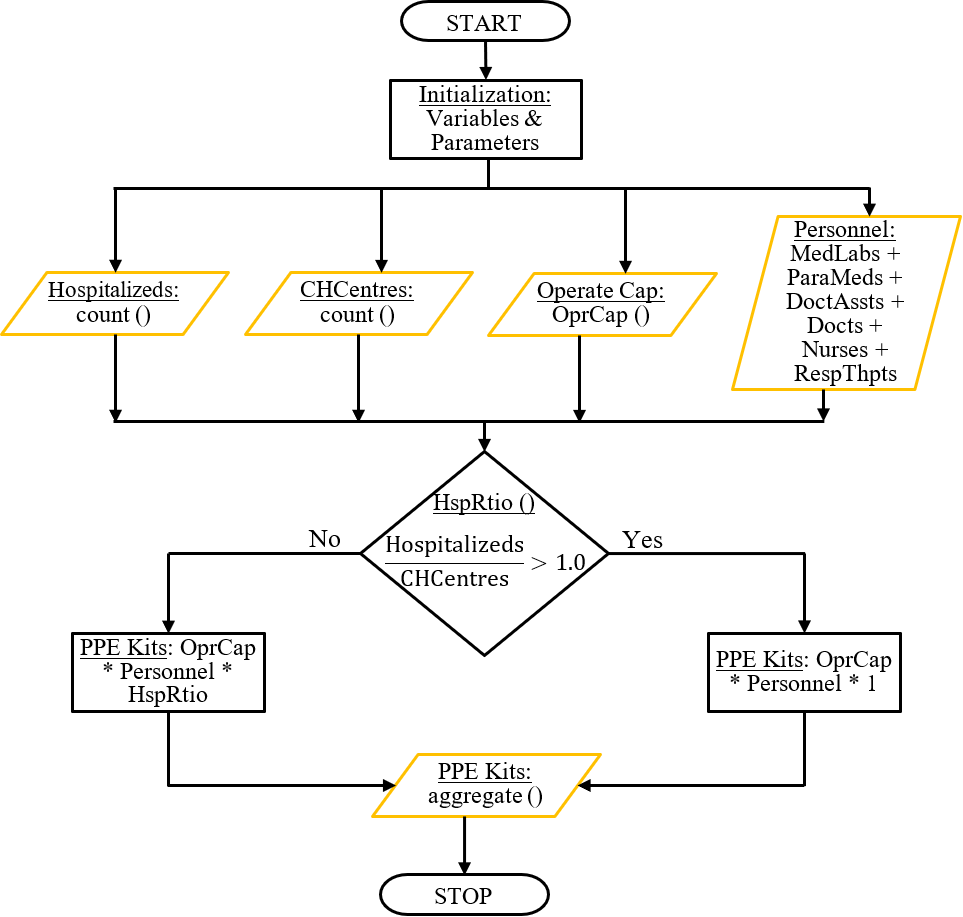}
\caption{Proposed architecture to predict \gls{ppe} demand in \gls{chc}.}
\label{Figure:ppe_model}
\end{figure}

Figure \ref{Figure:ppe_model} and Algorithm \ref{alg:ppe_model} showcase our proposed model with regard to the prediction of \gls{ppe} demand(s), by provincial \gls{chc}s, in relation to the \gls{covid19} pandemic. We have employed an interpolation technique herein, which relies on the predictions of hospitalized \gls{covid19} patients ($ y_{*,2} \subseteq Y^H $).

For each $ i^{th} $ day prediction of \gls{ppe} demand by a \gls{chc}, we instantiate Algorithm \ref{alg:ppe_model} and initialize the following variables:
\begin{enumerate} \label{ppe_init_variables}
\item $ i^{th} $ day prediction, \gls{hospital} \gls{covid19} cases ($ y_{i,2} $);
\item $ i^{th} $ day count of operational (regional) \gls{chc}s ($ feat\_11_i $);
\item $ i^{th} $ day average operating capacity of available \gls{covid19}-related workforce within regional \gls{chc}s ($ OprCap_i $);
\item The $ i^{th} $ day count of available frontline \gls{covid19}-related workforce ($ Personnel_i $) within regional \gls{chc}s.
\end{enumerate}

Subsequently, we compute the $ i^{th} $ day value for the dependent variable, $ HspRtio_i $, which denotes the average number of \gls{hospital} \gls{covid19} patients per \gls{chc} in each region. In other words, $ HspRtio_i = y_{i,2} : feat\_11_i $.

Afterwards, if $ HspRtio_i \geq 1 $ is true, it signifies that there exist at least $ 1 $ \gls{hospital} \gls{covid19} patient in every \gls{chc} for each region/province. Hence, we estimate the $ PPE\,Kits_i $ for the $ i^{th} $ day as the product: $ OprCap_i \times Personnel_i \times 1.0 $.

However, if $ HspRtio_i < 1 $ is true, it indicates that there exist some \gls{chc}s in each region with null \gls{hospital} \gls{covid19} patient. Therefore, we estimate the $ PPE\,Kits_i $ for that $ i^{th} $ day as the product: $ OprCap_i \times Personnel_i \times HspRtio_i $.

\section{Materials and Methods} \label{materials_methods}
Table \ref{Table:datasets} gives a detailed overview of the \gls{covid19} datasets (per Canadian province) employed herein for our experiments. We have implemented the following objective functions, with respect to benchmarking our proposed \gls{mtl} framework, namely: \gls{r2}, \gls{evs}, \gls{mae}, \gls{rmse}, and \gls{tt}. Also, to facilitate the implementation and evaluation of our work; we have used these libraries, viz: Scikit-Learn \cite{scikit-learn} and Keras \cite{Keras-Chollet}. The core regressor function of our proposed \gls{mtl} model is based on k-nearest neighbors \cite{scikit-learn}. The k-nearest-neighbors regressor has been implemented using its default hyperparameters, as in Scikit-Learn \cite{scikit-learn} library, with exception to the \emph{n\_neighbors} and \emph{weights} parameters which we have tuned as: \emph{KNeighborsRegressor(n\_neighbors=6, weights=`distance')}. Details regarding the reproducibility of our framework is available via: \url{https://github.com/bhevencious/COVID-19-Monitor/blob/main/README.md}

\begin{table}[h!]
\centering
\setlength{\tabcolsep}{0.3em}  
\renewcommand{\arraystretch}{0.9}  
\linespread{0.90}  
\rmfamily  
\caption{Benchmark datasets \cite{Berry2020, GoogleMobility2021, StatCan2021, CIHItline2021, CIHIdsets2021, ESRIdsets2021, GlobalNews2021}.}
     \begin{tabular}{r|p{5.5em}|p{3.0em}|p{3.0em}|p{16.0em}}  
     \hline
		 \multicolumn{1}{r|}{} & \multicolumn{1}{l|}{Dataset} & \multicolumn{1}{l|}{Start} & \multicolumn{1}{l|}{End} & \multicolumn{1}{l}{Description} \\
     \hline
		 $ 1 $ & Alberta & \multirow{5}{3.0em}{January $ 25^{th} $, 2020} & \multirow{12}{3.0em}{January $ 20^{th} $, 2021} & \multirow{7}{16.0em}{Each dataset, with regard to a province in Canada, contains variables of the target space and variables of the feature space. Every regional or provincial dataset comprises $ 362 $ rows which represent daily epidemiological records spanning from January $ 25^{th} $, 2020 to January $ 20^{th} $, 2021.} \\
		 $ 2 $ & British Columbia & & & \\
		 $ 3 $ & Manitoba & & & \\
		 $ 4 $ & New Brunswick & & & \\
		 $ 5 $ & Ontario & & & \\
		 $ 6 $ & Quebec & & & \\
		 $ 7 $ & Saskatchewan & & & \\
	 \hline
     \end{tabular}
\label{Table:datasets}
\end{table}

\section{Experiments, Results, and Discussions} \label{discussions}
\begin{figure}[h!]  
\centering
\includegraphics[width=0.495\textwidth]{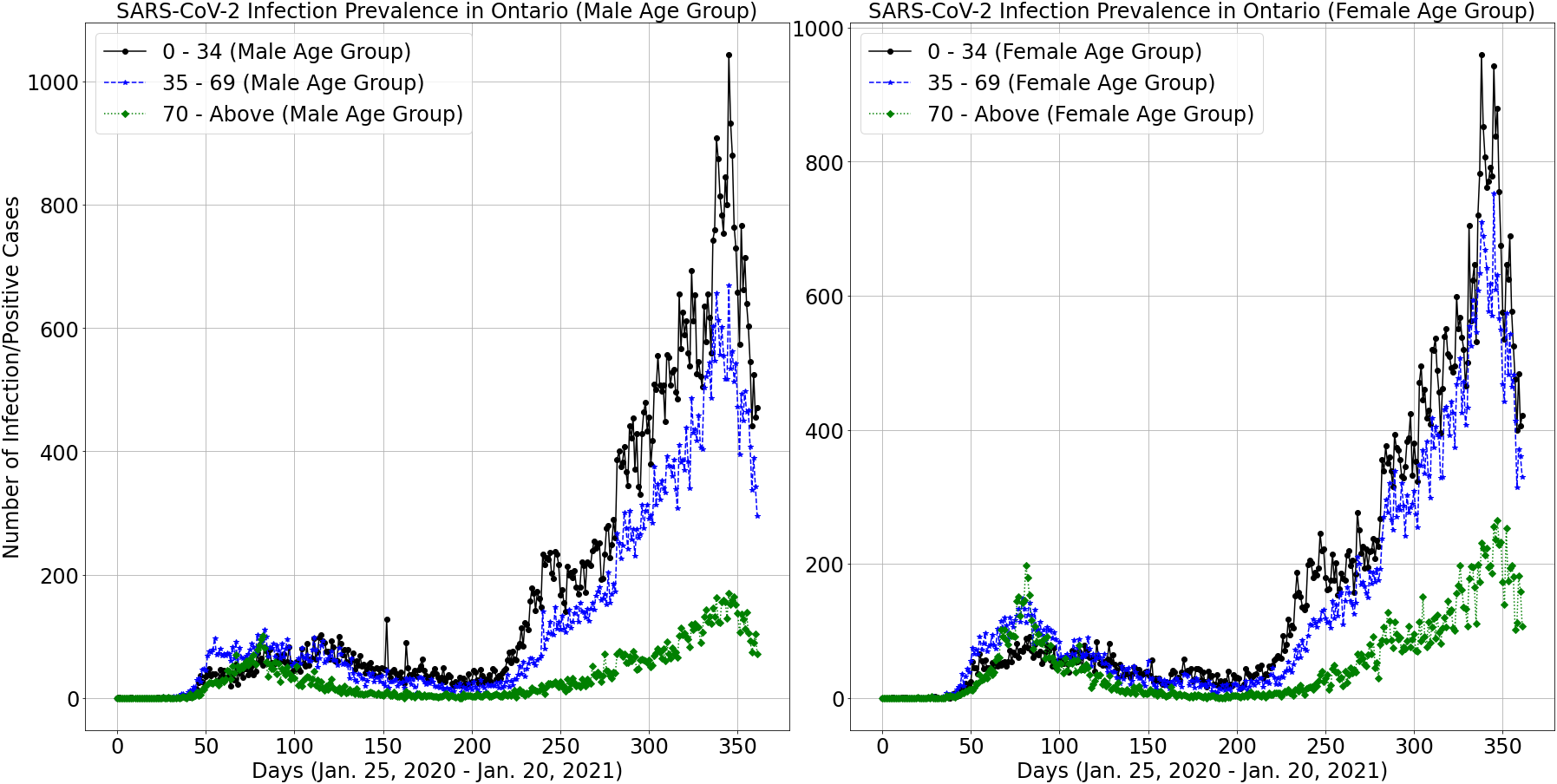}
\caption{\gls{covid19} prevalence across male and female age groups in Ontario, Canada.}
\label{Figure:infect_age_group}
\end{figure}

\begin{table*}[h!]
  \centering
  \setlength{\tabcolsep}{0.7em}  
  \renewcommand{\arraystretch}{0.9}  
  \caption{Experiment results for the prediction of \gls{infect} cases, based on 95\% Confidence Interval, on a test set comprising $ 54 $ randomly sampled days.}
    \begin{tabular}{|l|r|r|r|r|r|r|r|r|r|r|r|r|r|}
    \hline
    \multirow{2}{4.0em}{\textbf{Province}} & \multicolumn{3}{c|}{\textbf{\gls{r2}} (Prediction Intervals)} & \multicolumn{3}{c|}{\textbf{\gls{evs}} (Prediction Intervals)} & \multicolumn{3}{c|}{\textbf{\gls{mae}} (Prediction Intervals)} & \multicolumn{3}{c|}{\textbf{\gls{rmse}} (Prediction Intervals)} & \multirow{2}{*}{\textbf{\gls{tt}(s)}} \\
   	& \multicolumn{1}{c|}{\textbf{low}} & \multicolumn{1}{c|}{\textbf{mid}} & \multicolumn{1}{c|}{\textbf{top}} & \multicolumn{1}{c|}{\textbf{low}} & \multicolumn{1}{c|}{\textbf{mid}} & \multicolumn{1}{c|}{\textbf{top}} & \multicolumn{1}{c|}{\textbf{low}} & \multicolumn{1}{c|}{\textbf{mid}} & \multicolumn{1}{c|}{\textbf{top}} & \multicolumn{1}{c|}{\textbf{low}} & \multicolumn{1}{c|}{\textbf{mid}} & \multicolumn{1}{c|}{\textbf{top}} & \\
    \hline
	\textbf{Alberta} & 0.638 & 0.723 & 0.893 & 0.639 & 0.732 & 0.893 & 78.465 & 125.931 & 137.041 & 136.994 & 230.976 & 279.454 & 0.024 \\
	\textbf{British Columbia} & 0.028 & 0.473 & 0.851 & 0.038 & 0.485 & 0.854 & 60.238 & 102.674 & 123.090 & 134.696 & 238.968 & 267.852 & 0.007 \\
	\textbf{Manitoba} & 0.731 & 0.825 & 0.913 & 0.773 & 0.825 & 0.913 & 20.887 & 26.604 & 29.846 & 39.451 & 49.448 & 53.801 & 0.007 \\
	\textbf{New Brunswick} & 0.035 & 0.596 & 0.810 & 0.010 & 0.610 & 0.817 & 1.353 & 2.111 & 2.496 & 2.305 & 3.610 & 4.312 & 0.007 \\
	\textbf{Ontario} & 0.896 & 0.944 & 0.964 & 0.898 & 0.945 & 0.964 & 88.514 & 116.736 & 143.416 & 154.942 & 208.590 & 261.922 & 0.029 \\
	\textbf{Quebec} & 0.876 & 0.876 & 0.932 & 0.878 & 0.885 & 0.932 & 107.571 & 116.521 & 141.068 & 162.399 & 176.061 & 205.731 & 0.007 \\
	\textbf{Saskatchewan} & 0.676 & 0.805 & 0.883 & 0.686 & 0.813 & 0.884 & 16.755 & 21.284 & 26.352 & 27.744 & 39.989 & 50.375 & 0.008 \\
    \hline
    \end{tabular}%
  \label{Table:tl_results_1}%
\end{table*}%

\begin{table*}[h!]
  \centering
  \setlength{\tabcolsep}{0.7em}  
  \renewcommand{\arraystretch}{0.9}  
  \caption{Experiment results for the prediction of \gls{hospital} cases, based on 95\% Confidence Interval, on a test set comprising $ 54 $ randomly sampled days.}
    \begin{tabular}{|l|r|r|r|r|r|r|r|r|r|r|r|r|r|}
    \hline
    \multirow{2}{4.0em}{\textbf{Province}} & \multicolumn{3}{c|}{\textbf{\gls{r2}} (Prediction Intervals)} & \multicolumn{3}{c|}{\textbf{\gls{evs}} (Prediction Intervals)} & \multicolumn{3}{c|}{\textbf{\gls{mae}} (Prediction Intervals)} & \multicolumn{3}{c|}{\textbf{\gls{rmse}} (Prediction Intervals)} & \multirow{2}{*}{\textbf{\gls{tt}(s)}} \\
   	& \multicolumn{1}{c|}{\textbf{low}} & \multicolumn{1}{c|}{\textbf{mid}} & \multicolumn{1}{c|}{\textbf{top}} & \multicolumn{1}{c|}{\textbf{low}} & \multicolumn{1}{c|}{\textbf{mid}} & \multicolumn{1}{c|}{\textbf{top}} & \multicolumn{1}{c|}{\textbf{low}} & \multicolumn{1}{c|}{\textbf{mid}} & \multicolumn{1}{c|}{\textbf{top}} & \multicolumn{1}{c|}{\textbf{low}} & \multicolumn{1}{c|}{\textbf{mid}} & \multicolumn{1}{c|}{\textbf{top}} & \\
    \hline
	\textbf{Alberta} & 0.927 & 0.942 & 0.968 & 0.930 & 0.942 & 0.969 & 17.324 & 25.341 & 27.724 & 30.250 & 51.036 & 60.819 & 0.024 \\
	\textbf{British Columbia} & 0.296 & 0.944 & 0.983 & 0.253 & 0.945 & 0.983 & 10.588 & 16.465 & 38.972 & 17.351 & 25.163 & 139.776 & 0.007 \\
	\textbf{Manitoba} & 0.924 & 0.954 & 0.989 & 0.931 & 0.955 & 0.989 & 6.954 & 12.120 & 12.622 & 12.423 & 26.279 & 32.651 & 0.007 \\
	\textbf{New Brunswick} & 0.035 & 0.500 & 0.719 & 0.161 & 0.520 & 0.734 & 0.482 & 0.704 & 0.935 & 0.934 & 1.232 & 1.539 & 0.007 \\
	\textbf{Ontario} & 0.881 & 0.924 & 0.970 & 0.884 & 0.924 & 0.974 & 41.163 & 69.221 & 85.747 & 65.841 & 112.323 & 145.947 & 0.029 \\
	\textbf{Quebec} & 0.739 & 0.742 & 0.777 & 0.741 & 0.746 & 0.780 & 137.042 & 146.792 & 160.369 & 258.573 & 279.158 & 284.480 & 0.007 \\
	\textbf{Saskatchewan} & 0.802 & 0.948 & 0.970 & 0.830 & 0.949 & 0.972 & 5.197 & 6.208 & 9.201 & 8.742 & 11.618 & 18.474 & 0.008 \\
    \hline
    \end{tabular}%
  \label{Table:tl_results_2}%
\end{table*}%

\begin{table*}[h!]
  \centering
  \setlength{\tabcolsep}{0.7em}  
  \renewcommand{\arraystretch}{0.9}  
  \caption{Experiment results for the prediction of \gls{recover} cases, based on 95\% Confidence Interval, on a test set comprising $ 54 $ randomly sampled days.}
    \begin{tabular}{|l|r|r|r|r|r|r|r|r|r|r|r|r|r|}
    \hline
    \multirow{2}{4.0em}{\textbf{Province}} & \multicolumn{3}{c|}{\textbf{\gls{r2}} (Prediction Intervals)} & \multicolumn{3}{c|}{\textbf{\gls{evs}} (Prediction Intervals)} & \multicolumn{3}{c|}{\textbf{\gls{mae}} (Prediction Intervals)} & \multicolumn{3}{c|}{\textbf{\gls{rmse}} (Prediction Intervals)} & \multirow{2}{*}{\textbf{\gls{tt}(s)}} \\
   	& \multicolumn{1}{c|}{\textbf{low}} & \multicolumn{1}{c|}{\textbf{mid}} & \multicolumn{1}{c|}{\textbf{top}} & \multicolumn{1}{c|}{\textbf{low}} & \multicolumn{1}{c|}{\textbf{mid}} & \multicolumn{1}{c|}{\textbf{top}} & \multicolumn{1}{c|}{\textbf{low}} & \multicolumn{1}{c|}{\textbf{mid}} & \multicolumn{1}{c|}{\textbf{top}} & \multicolumn{1}{c|}{\textbf{low}} & \multicolumn{1}{c|}{\textbf{mid}} & \multicolumn{1}{c|}{\textbf{top}} & \\
    \hline
	\textbf{Alberta} & 0.550 & 0.674 & 0.909 & 0.555 & 0.680 & 0.909 & 79.439 & 84.789 & 127.683 & 138.549 & 162.563 & 243.923 & 0.024 \\
	\textbf{British Columbia} & 0.100 & 0.366 & 0.798 & 0.031 & 0.384 & 0.804 & 50.762 & 92.780 & 137.169 & 119.584 & 238.078 & 321.961 & 0.007 \\
	\textbf{Manitoba} & 0.069 & 0.155 & 0.812 & 0.084 & 0.179 & 0.817 & 24.554 & 51.337 & 105.441 & 47.630 & 188.434 & 552.707 & 0.007 \\
	\textbf{New Brunswick} & 0.076 & 0.237 & 0.557 & 0.012 & 0.238 & 0.562 & 1.143 & 1.776 & 2.300 & 2.240 & 2.931 & 3.762 & 0.007 \\
	\textbf{Ontario} & 0.894 & 0.946 & 0.958 & 0.899 & 0.948 & 0.961 & 74.630 & 106.374 & 130.069 & 124.021 & 183.136 & 235.731 & 0.029 \\
	\textbf{Quebec} & 0.161 & 0.521 & 0.723 & 0.020 & 0.460 & 0.752 & 187.252 & 233.452 & 304.964 & 428.358 & 739.132 & 1057.534 & 0.007 \\
	\textbf{Saskatchewan} & 0.387 & 0.546 & 0.734 & 0.026 & 0.451 & 0.736 & 17.187 & 25.794 & 37.825 & 43.580 & 59.529 & 101.768 & 0.008 \\
    \hline
    \end{tabular}%
  \label{Table:tl_results_3}%
\end{table*}%

\begin{table*}[h!]
  \centering
  \setlength{\tabcolsep}{0.7em}  
  \renewcommand{\arraystretch}{0.9}  
  \caption{Experiment results for the prediction of \gls{death} cases, based on 95\% Confidence Interval, on a test set comprising $ 54 $ randomly sampled days.}
    \begin{tabular}{|l|r|r|r|r|r|r|r|r|r|r|r|r|r|}
    \hline
    \multirow{2}{4.0em}{\textbf{Province}} & \multicolumn{3}{c|}{\textbf{\gls{r2}} (Prediction Intervals)} & \multicolumn{3}{c|}{\textbf{\gls{evs}} (Prediction Intervals)} & \multicolumn{3}{c|}{\textbf{\gls{mae}} (Prediction Intervals)} & \multicolumn{3}{c|}{\textbf{\gls{rmse}} (Prediction Intervals)} & \multirow{2}{*}{\textbf{\gls{tt}(s)}} \\
   	& \multicolumn{1}{c|}{\textbf{low}} & \multicolumn{1}{c|}{\textbf{mid}} & \multicolumn{1}{c|}{\textbf{top}} & \multicolumn{1}{c|}{\textbf{low}} & \multicolumn{1}{c|}{\textbf{mid}} & \multicolumn{1}{c|}{\textbf{top}} & \multicolumn{1}{c|}{\textbf{low}} & \multicolumn{1}{c|}{\textbf{mid}} & \multicolumn{1}{c|}{\textbf{top}} & \multicolumn{1}{c|}{\textbf{low}} & \multicolumn{1}{c|}{\textbf{mid}} & \multicolumn{1}{c|}{\textbf{top}} & \\
    \hline
	\textbf{Alberta} & 0.256 & 0.531 & 0.808 & 0.272 & 0.538 & 0.812 & 1.304 & 1.870 & 1.996 & 2.181 & 3.715 & 4.186 & 0.024 \\
	\textbf{British Columbia} & 0.037 & 0.352 & 0.605 & 0.076 & 0.301 & 0.607 & 1.310 & 2.412 & 4.125 & 2.823 & 5.599 & 10.616 & 0.007 \\
	\textbf{Manitoba} & 0.609 & 0.692 & 0.908 & 0.637 & 0.716 & 0.908 & 0.556 & 0.906 & 1.183 & 1.234 & 2.084 & 2.378 & 0.007 \\
	\textbf{New Brunswick} & 0.056 & 0.588 & 1.000 & 0.000 & 0.563 & 1.000 & 0.000 & 0.055 & 0.093 & 0.000 & 0.235 & 0.346 & 0.007 \\
	\textbf{Ontario} & 0.438 & 0.606 & 0.692 & 0.443 & 0.606 & 0.697 & 5.330 & 7.493 & 8.533 & 8.739 & 12.386 & 16.231 & 0.029 \\
	\textbf{Quebec} & 0.331 & 0.394 & 0.733 & 0.364 & 0.434 & 0.740 & 10.097 & 11.847 & 13.774 & 16.653 & 22.410 & 26.058 & 0.007 \\
	\textbf{Saskatchewan} & 0.085 & 0.256 & 0.600 & 0.013 & 0.266 & 0.608 & 0.301 & 0.434 & 0.707 & 0.736 & 1.139 & 1.581 & 0.008 \\
    \hline
    \end{tabular}%
  \label{Table:tl_results_4}%
\end{table*}%

In our experiments and results stated herein (Tables \ref{Table:tl_results_1}, \ref{Table:tl_results_2}, \ref{Table:tl_results_3}, and \ref{Table:tl_results_4}), we have rotated the training dataset for the \emph{Dedicated} \gls{ml}-model component of our \gls{mtl} framework across seven (7) distinct Canadian provinces (Alberta, British Columbia, Manitoba, New Brunswick, Ontario, Quebec, and Saskatchewan). Thus, any region/province used for training the \emph{Dedicated} \gls{ml}-model component (of our \gls{mtl} framework) is excluded from the training datasets for the \emph{Generic} \gls{ml}-model component of our \gls{mtl} framework. The objective functions (\gls{r2}, \gls{evs}) and (\gls{mae}, \gls{rmse}, \gls{tt}) attain their best at the values of $ 1.0 $ and $ 0.0 $, respectively.

Considering our experiment results in Table \ref{Table:tl_results_1}, Table \ref{Table:tl_results_2}, Table \ref{Table:tl_results_3}, and Table \ref{Table:tl_results_4}; the prediction interval of our proposed \gls{mtl} framework is based on a 95\% confidence interval. Thus, as the respective values for \gls{r2} approach unity ($ 1 $), they signify that our proposed \gls{mtl} framework fits adequately to the \gls{covid19} dataset(s); and otherwise, as \gls{r2} approaches zero ($ 0 $). In a similar fashion, as the respective values for \gls{evs} tend to unity ($ 1 $), they explain to us that our proposed \gls{mtl} model effectively captures and utilizes the data-point variations in the \gls{covid19} dataset(s). \gls{evs} explains otherwise as its respective values tend to zero ($ 0 $). Moreover, \gls{mae} and \gls{rmse} compare the predictions from our proposed \gls{mtl} model and the ground truth. As the respective values for \gls{mae} and \gls{rmse} approach zero ($ 0 $), they imply that our \gls{mtl} model makes predictions with relatively lower residual error(s).

Algorithm \ref{alg:ppe_model} proposed herein effects the prediction for \gls{ppe}-kits via interpolation into the predictions for \gls{hospital} cases. In consideration of the feature selection process carried out herein in subsection \ref{layer_3}, we have observed that human-related factors (precisely peer groups) most significantly influence the rates of \gls{infect} cases, \gls{hospital} cases, and \gls{death} cases as can be seen from Table \ref{Table:relv_feats_space}. Also, government actions (such as inoculation, pandemic wave, etc.) most significantly influence the rate of \gls{recover} cases as shown in Table \ref{Table:relv_feats_space}.

Figure \ref{Figure:infect_age_group} represents a distribution plot of \gls{covid19} prevalence across three major age groups in the province of Ontario (Canada). On one hand, we can see that youths (males and females whom fall within the age group of 0 to 34) greatly influence the spread of \gls{sars-cov-2} within a given socially interacting populace. On the other hand, seniors (males and females whom are of age 70 and above) are less likely to influence the spread of \gls{sars-cov-2} within a given socially interacting populace. However, these seniors (age 70 and above) remain the most susceptible to \gls{sars-cov-2} due to several age-related risk factors.

A known limitation of this work is that the datasets employed herein for our experiments and analyses were gathered from the date range of January $ 25^{th} $, 2020 to January $ 20^{th} $, 2021. Hence, we assumed that these datasets represent and reflect casualties or cases with reference to the earliest variant of \gls{sars-cov-2} (\gls{uk_variant}). Consecutively, our assumption herein with regard to each unit of \gls{ppe}-kit is that it comprises five (5) items, namely: a face shield, a N95 respirator or facemask, a pair of hand gloves, a pair of shoe covers, and an overall isolation gown.

\section{Acknowledgement} \label{acknowledgement}
This case has been funded by the Canadian Institutes of Health Research Operating Grant: COVID-19 May 2020 Rapid Research Funding Opportunity [operating grant VR5 172669]. Resources used in preparing this research were provided by the Province of Ontario, the Government of Canada through CIFAR, and companies sponsoring the Vector Institute.

\bibliographystyle{IEEEtran}
\bibliography{MOLOKWU_SHUVO_KOBTI_SNOWDON_ICTS4eHealth_2021}

\begin{thebibliography}{10}
\providecommand{\url}[1]{#1}
\csname url@samestyle\endcsname
\providecommand{\newblock}{\relax}
\providecommand{\bibinfo}[2]{#2}
\providecommand{\BIBentrySTDinterwordspacing}{\spaceskip=0pt\relax}
\providecommand{\BIBentryALTinterwordstretchfactor}{4}
\providecommand{\BIBentryALTinterwordspacing}{\spaceskip=\fontdimen2\font plus
\BIBentryALTinterwordstretchfactor\fontdimen3\font minus
  \fontdimen4\font\relax}
\providecommand{\BIBforeignlanguage}[2]{{%
\expandafter\ifx\csname l@#1\endcsname\relax
\typeout{** WARNING: IEEEtran.bst: No hyphenation pattern has been}%
\typeout{** loaded for the language `#1'. Using the pattern for}%
\typeout{** the default language instead.}%
\else
\language=\csname l@#1\endcsname
\fi
#2}}
\providecommand{\BIBdecl}{\relax}
\BIBdecl

\bibitem{WHOCD2020}
W.~H. Organization, ``Coronavirus disease (covid-19) dashboard,'' \emph{World
  Health Organization (WHO).}, 2020.

\bibitem{BenShlomo2002}
Y.~Ben-Shlomo and D.~Kuh, ``A life course approach to chronic disease
  epidemiology: conceptual models, empirical challenges and interdisciplinary
  perspectives.'' \emph{International Journal of Epidemiology.}, vol. 31 2, pp.
  285--293, 2004.

\bibitem{Burr2016}
H.~Burr, M.~Formazin, and A.~Pohrt, ``Methodological and conceptual issues
  regarding occupational psychosocial coronary heart disease epidemiology.''
  \emph{Scandinavian Journal of Work, Environment and Health.}, vol. 42 3, pp.
  251--255, 2016.

\bibitem{Bernasconi2020}
A.~Bernasconi, A.~Canakoglu, P.~Pinoli, and S.~Ceri, ``Empowering virus
  sequences research through conceptual modeling,'' \emph{bioRxiv.}, 2020.

\bibitem{Lin2020ACM}
Q.~Lin, S.~Zhao, D.~Gao, Y.~Lou, S.~Yang, S.~S. Musa, M.~Wang, Y.~Cai, W.~Wang,
  L.~Yang, and D.~He, ``A conceptual model for the coronavirus disease 2019
  (covid-19) outbreak in wuhan, china with individual reaction and governmental
  action,'' \emph{International Journal of Infectious Diseases.}, vol.~93, pp.
  211 -- 216, 2020.

\bibitem{Brauer2008Compartmentals}
F.~Brauer, ``Compartmental models in epidemiology,'' \emph{Mathematical
  Epidemiology}, vol. 1945, pp. 19 -- 79, 2008.

\bibitem{Hethcote2000}
H.~Hethcote, ``The mathematics of infectious diseases,'' \emph{SIAM Review.},
  vol.~42, pp. 599--653, 2000.

\bibitem{Zhao2020Modeling}
S.~Zhao and H.~Chen, ``Modeling the epidemic dynamics and control of covid-19
  outbreak in china,'' \emph{Quantitative Biology (Beijing, China).}, pp. 1 --
  9, 2020.

\bibitem{Aleta2020}
A.~Aleta, D.~Mart{\'i}n-Corral, A.~P. y~Piontti, M.~Ajelli, M.~Litvinova,
  M.~Chinazzi, N.~Dean, M.~Halloran, I.~Longini, S.~Merler, A.~Pentland,
  A.~Vespignani, E.~Moro, and Y.~Moreno, ``Modeling the impact of social
  distancing, testing, contact tracing and household quarantine on second-wave
  scenarios of the covid-19 epidemic,'' \emph{medRxiv.}, 2020.

\bibitem{Rockett2020}
R.~Rockett, A.~Arnott, C.~Lam, R.~Sadsad, V.~J. Timms, K.-A. Gray, J.~Eden,
  S.~Chang, M.~Gall, J.~Draper, E.~Sim, N.~Bachmann, I.~Carter, K.~Basile,
  R.~Byun, M.~O'Sullivan, S.~C. Chen, S.~Maddocks, T.~C. Sorrell, D.~Dwyer,
  E.~Holmes, J.~Kok, M.~Prokopenko, and V.~Sintchenko, ``Revealing covid-19
  transmission in australia by sars-cov-2 genome sequencing and agent-based
  modeling,'' \emph{Nature Medicine.}, pp. 1 -- 7, 2020.

\bibitem{SILVA2020110088}
P.~C. Silva, P.~V. Batista, H.~S. Lima, M.~A. Alves, F.~G. Guimarães, and
  R.~C. Silva, ``Covid-abs: An agent-based model of covid-19 epidemic to
  simulate health and economic effects of social distancing interventions,''
  \emph{Chaos, Solitons \& Fractals.}, vol. 139, p. 110088, 2020.

\bibitem{Shuvo2020S}
S.~B. Shuvo, B.~C. Molokwu, and Z.~Kobti, ``Simulating the impact of hospital
  capacity and social isolation to minimize the propagation of infectious
  diseases,'' in \emph{Proceedings of the 26th ACM SIGKDD International
  Conference on Knowledge Discovery \& Data Mining.}, New York, NY, United
  States, 2020.

\bibitem{Zou2020}
D.~Zou, L.~Wang, P.~Xu, J.~Chen, W.~Zhang, and Q.~Gu, ``Epidemic model guided
  machine learning for covid-19 forecasts in the united states,''
  \emph{medRxiv.}, 2020.

\bibitem{Kukar2020}
M.~Kukar, G.~Guncar, T.~Vovko, S.~Podnar, P.~Cernelc, M.~Brvar, M.~Zalaznik,
  M.~Notar, S.~Moskon, and M.~Notar, ``Covid-19 diagnosis by routine blood
  tests using machine learning,'' \emph{ArXiv, Physics}, 2020.

\bibitem{Berry2020}
I.~Berry, J.-P.~R. Soucy, A.~Tuite, and D.~Fisman, ``Open access epidemiologic
  data and an interactive dashboard to monitor the covid-19 outbreak in
  canada,'' \emph{Canadian Medical Association Journal.}, vol. 192, pp. E420 --
  E420, 2020.

\bibitem{GoogleMobility2021}
{Google LLC}, ``Google covid-19 community mobility reports,''
  \url{https://www.google.com/covid19/mobility/}, Mountain View, CA, 2021.

\bibitem{StatCan2021}
{Statistics Canada}, ``Covid-19 statcan covid-19: Data to insights for a better
  canada [data tables],''
  \url{https://www150.statcan.gc.ca/n1/en/catalogue/45280001}, Ottawa, Canada,
  2021.

\bibitem{scikit-learn}
F.~Pedregosa, G.~Varoquaux, A.~Gramfort, V.~Michel, B.~Thirion, O.~Grisel,
  M.~Blondel, P.~Prettenhofer, R.~Weiss, V.~Dubourg, J.~Vanderplas, A.~Passos,
  D.~Cournapeau, M.~Brucher, M.~Perrot, and E.~Duchesnay, ``Scikit-learn:
  Machine learning in {P}ython,'' \emph{Journal of Machine Learning Research.},
  vol.~12, pp. 2825--2830, 2011.

\bibitem{AlanJeff2007}
I.~S. Gradshteyn and I.~M. Ryzhik, \emph{Table of Integrals, Series, and
  Products}.\hskip 1em plus 0.5em minus 0.4em\relax San Diego, CA: Academic
  Press; 7th edition, 2007, ch. 15. Norms.

\bibitem{Keras-Chollet}
F.~Chollet, Ed., \emph{Deep Learning with Python}.\hskip 1em plus 0.5em minus
  0.4em\relax Shelter Island, NY: Manning Publications, 2017.

\bibitem{CIHItline2021}
{Canadian Institute for Health Information}, ``Covid-19 intervention timeline
  in canada,''
  \url{https://www.cihi.ca/en/covid-19-intervention-timeline-in-canada},
  Ottawa, Canada, 2021.

\bibitem{CIHIdsets2021}
------, ``Access data and reports,''
  \url{https://www.cihi.ca/en/access-data-and-reports}, Ottawa, Canada, 2021.

\bibitem{ESRIdsets2021}
{Esri Canada}, ``Covid-19 open data,''
  \url{https://resources-covid19canada.hub.arcgis.com/pages/open-data},
  Toronto, Canada, 2021.

\bibitem{GlobalNews2021}
{Global News}, ``Coronavirus,'' \url{https://globalnews.ca/tag/coronavirus/},
  Toronto, Canada, 2021.

\end{thebibliography}

\end{document}